\documentclass[11pt,a4paper]{article}
\usepackage{authblk}
\usepackage[hyperref]{acl2019}
\usepackage{times}
\usepackage{latexsym}
\usepackage{booktabs}
\usepackage{multirow}
\usepackage{tikz}
\usepackage{tikz-dependency}
\usepackage{todonotes}
\usepackage{url}
\aclfinalcopy 
\def\aclpaperid{1296} 

\title{Evaluating Discourse in Structured Text Representations}
\author[1]{{\bf Elisa Ferracane}}
\author[2]{{\bf Greg Durrett}}
\author[1]{{\bf Junyi Jessy Li}}
\author[1]{{\bf Katrin Erk}}
\affil[1]{Department of Linguistics}
\affil[2]{Department of Computer Science}
\affil[ ]{The University of Texas at Austin}
\affil[ ]{\tt elisa@ferracane.com, gdurrett@cs.utexas.edu}
\affil[ ]{\tt jessy@austin.utexas.edu, katrin.erk@mail.utexas.edu}

\date{}

\begin{document}
\maketitle
\begin{abstract}
Discourse structure is integral to understanding a text and is helpful in many NLP tasks. Learning \emph{latent} representations of discourse is an attractive alternative to acquiring expensive labeled discourse data.
\citet{Liu:2018} propose a structured attention mechanism for text classification that derives a tree over a text, akin to an RST discourse tree. We examine this model in detail, and evaluate on additional discourse-relevant tasks and datasets, in order to assess whether the structured attention improves performance on the end task and whether it captures a text's discourse structure. We find the learned latent trees have little to no structure and instead focus on lexical cues; even after obtaining more structured trees with proposed model modifications, the trees are still far from capturing discourse structure when compared to discourse dependency trees from an existing discourse parser. Finally, ablation studies show the structured attention provides little benefit, sometimes even hurting performance.\footnote{Code and data available at \url{https://github.com/elisaF/structured}}
\end{abstract}

\section{Introduction}
Discourse describes how a document is organized, and how discourse units are rhetorically connected to each other. 
Taking into account this structure has shown to help many NLP end tasks, including summarization \cite{Hirao:2013,Durrett:2016}, machine translation \cite{Joty:2017}, and sentiment analysis \cite{Ji:2017}. However, annotating discourse requires considerable effort by trained experts and may not always yield a structure appropriate for the end task. As a result, having a model induce the discourse structure of a text is an attractive option. Our goal in this paper is to evaluate such an induced structure. 

Inducing structure has been a recent popular approach in syntax \cite{Yogatama:2017,Choi:2018,Bisk:2018}. Evaluations of these latent trees have shown they are inconsistent, shallower than their explicitly parsed counterparts (Penn Treebank parses) and do not resemble any linguistic syntax theory \cite{Williams:2018}.

For discourse, \citet{Liu:2018} (L\&L) induce a document-level structure while performing text classification with a structured attention that is constrained to resolve to a non-projective dependency tree. We evaluate the document-level structure induced by this model. In order to compare the induced structure to existing linguistically-motivated structures, we choose Rhetorical Structure Theory (RST) \cite{Mann:1988}, a widely-used framework for discourse structure, because it also produces tree-shaped structures.\footnote{The Penn Discourse Treebank \citep[PDTB;][]{Prasad:2008} captures lexically-grounded discourse for individual connectives and adjacent sentences, and does not span an entire document; Segmented Discourse Representation Theory \cite{Lascarides:2008} is a graph.}
 We evaluate on some of the same tasks as L\&L, but add two more tasks we theorize to be more discourse-sensitive: text classification of writing quality, and sentence order discrimination (as proposed by \citet{Barzilay:2008}). 

Our research uncovers multiple negative results. We find that, contrary to L\&L, the structured attention does not help performance in most cases; further, the model is not learning discourse. Instead, the model learns trees with little to no structure heavily influenced by lexical cues to the task. In an effort to induce better trees, we propose several principled modifications to the model, some of which yield more structured trees. However, even the more structured trees bear little resemblance to ground truth RST trees.

We conclude the model holds promise, but requires moving beyond text classification, and injecting supervision (as in \citet{Strubell:2018}). 


Our contributions are (1) comprehensive performance results on existing and additional tasks and datasets showing document-level structured attention is largely unhelpful, (2) in-depth analyses of induced trees showing they do not represent discourse, and (3) several principled model changes to produce better structures but that still do not resemble the structure of discourse.

\section{Rhetorical Structure Theory (RST)}
In RST, coherent texts consist of minimal units, which are linked to each other, recursively, through rhetorical relations \cite{Mann:1988}. Thus, the goal of RST is to describe the rhetorical organization of a text by using a hierarchical tree structure that captures the communicative intent of the writer. An RST discourse tree can further be represented as a \emph{discourse dependency tree}. We follow the algorithm of \citet{Hirao:2013} to create an unlabelled dependency tree based on the nuclearity of the tree.

\section{Models}
\label{sec:models}
We present two models: one for text classification, and one for sentence ordering. Both are based on the L\&L model, with a design change to cause stronger percolation of information up the tree (we also experiment without this change). 


\begin{figure}[t]
\centering
\includegraphics[width=0.48\textwidth,height=30mm]{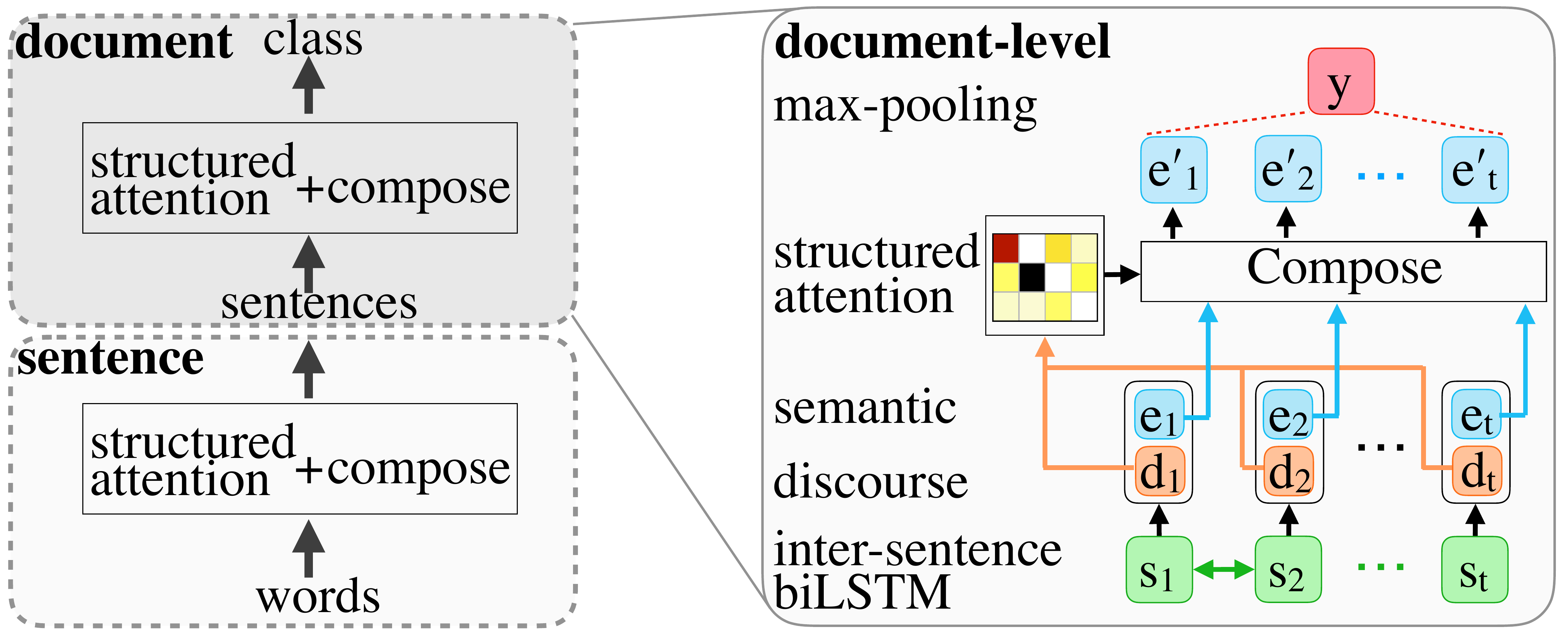}
\vspace{-1.9em}
\caption{Model of \citet{Liu:2018} with the document-level portion (right) that composes sentences into a document representation.}
\label{fig:model}
\vspace{-0.9em}
\end{figure}

\begin{table*}[th!]
\small
\centering
\scalebox{0.93}{
\begin{tabular}{llllll}
\toprule
    & Yelp & Debates & WQ & WQTC & WSJSO \\
\midrule
L\&L(orig) &68.51\thinspace$|$\thinspace\ \textbf{68.27} (0.19) &81.82\thinspace$|$\thinspace\ 79.48 (2.90) &84.14\thinspace$|$\thinspace\ \textbf{82.69} (1.36) &80.73\thinspace$|$\thinspace\ 79.63 (1.03) &96.17\thinspace$|$\thinspace\ \textbf{95.29} (0.84) \\
L\&L(ours)           & 68.51\thinspace$|$\thinspace\ \textbf{68.23} (0.23)  & 78.88\thinspace$|$\thinspace\ 77.81 (1.80)  & 84.14\thinspace$|$\thinspace\ \textbf{82.70} (1.36)  & 82.49\thinspace$|$\thinspace\ \textbf{81.11} (0.95) & 95.57\thinspace$|$\thinspace\ \textbf{94.76 }(1.11) \\ 
$-$doc attn & 68.34\thinspace$|$\thinspace\ \textbf{68.13} (0.17) & 82.89\thinspace$|$\thinspace\ \textbf{81.42} (1.08)  & 83.75\thinspace$|$\thinspace\ \textbf{82.80} (0.94)    & 80.60\thinspace$|$\thinspace\ 79.25 (0.94) & 95.57\thinspace$|$\thinspace\ \textbf{95.11} (0.42) \\
$-$both attn & 68.19\thinspace$|$\thinspace\ 68.05 (0.13) & 79.95\thinspace$|$\thinspace\ 77.34 (1.79)  & 84.27\thinspace$|$\thinspace\ \textbf{83.16} (1.25)  & 77.58\thinspace$|$\thinspace\ 76.16 (1.25) & 95.23\thinspace$|$\thinspace\ \textbf{94.68} (0.37)       \\
\midrule
L\&L(reported) & 68.6 & 76.5  & - & - & - \\
\bottomrule
\end{tabular}
}
\vspace{-0.7em}
\caption{Max\thinspace$|$\thinspace\ mean (standard deviation) accuracy on the test set averaged across four training runs with different initialization weights. Bolded numbers are within 1 standard deviation of the best performing model. L\&L(orig) uses the original L\&L code; L\&L(ours) includes the design change and bug fix. L\&L(reported) lists results reported by L\&L on a single training run.
}
\label{tab:results_full}
\vspace{-1.5em}
\end{table*}

\smallskip
\noindent \textbf{Text classification} The left-hand side of Figure \ref{fig:model} presents an overview of the model: the model operates first at the sentence-level to create sentence representations, and then at the document-level to create a document representation from the previously created sentence representations. In more detail, the model composes GloVe embeddings \cite{Pennington:2014} into a sentence representation using structured attention (from which a tree can be derived), then sentence representations into a single document representation for class prediction. At both sentence and document level, each object (word or sentence, respectively) attends to other objects that could be its parent in the tree. Since the sentence and document-level parts of the model are identical, we focus on the document level (Figure \ref{fig:model}, right), which is of interest to us for evaluating discourse effects. 

Sentence representations $s_1, \ldots, s_t$ are fed to a bidirectional LSTM, and the hidden representations $[h_1 , \ldots, h_t]$ consist of a semantic part ($e_t$) and a structure part ($d_t$):
$[\mathbf{e}_t, \mathbf{d}_t] = \mathbf{h}_t$.
Unnormalized scores $\mathbf{f}_{ij}$ representing potentials between parent $i$ and child $j$ are calculated using a bilinear function over the structure vector:
\vspace{-0.7em}
\begin{equation}
\mathbf{t}_p = \tanh(\mathbf{W}_p\mathbf{d}_i);\hspace{20pt} \mathbf{t}_c = \tanh(\mathbf{W}_c\mathbf{d}_j)
\vspace{-0.7em}
\end{equation} 
\begin{equation}
\mathbf{f}_{ij} = \mathbf{t}^T_p \mathbf{W}_a\mathbf{t}_c
\vspace{-0.3em}
\end{equation}
The matrix-tree theorem allows us to compute marginal probabilities $a_{ij}$ of dependency arcs under the distribution over non-projective dependency trees induced by $\mathbf{f}_{ij}$ (details in \citet{Koo:2007}). This computation is fully differentiable, allowing it to be treated as another neural network layer in the model. We importantly note the model only uses the \emph{marginals}. We can post-hoc use the Chu-Liu-Edmonds algorithm to retrieve the highest-scoring tree under $\mathbf{f}$, which we call $\mathbf{f_{best}}$ \cite{Chu:1965,Edmonds:1967}.

The semantic vectors of sentences $\mathbf{e}$ are then updated using this attention. Here we diverge from the L\&L model: in their implementation,\footnote{\url{https://github.com/nlpyang/structured}} each node is updated based on a weighted sum over its \emph{parents} in the tree (their paper states both parents and children).\footnote{We found similar results for using both parents and children as well as using parents only.} We instead inform each node by a sum over its \emph{children}, more in line with past work where information more intuitively percolates from children to parents and not the other way \cite{Ji:2017} (we also run experiments without this design change).
We calculate the context for all possible children of that sentence as: 
\vspace{-0.7em}
\begin{equation}
\label{eq:children}
c_i = \sum^{n}_{k=1}a_{ik}e_{k}
\vspace{-0.3em}
\end{equation}
where $a_{ik}$ is the probability that $k$ is the child of $i$, and $e_k$ is the semantic vector of the child.

The children vectors are then passed through a non-linear function, resulting in the \emph{updated} semantic vector $e'_i$ for parent node $i$.
\vspace{-0.7em}
\begin{equation}
\label{eq:update}
e'_i = \tanh(W_r[e_i, c_i])
\vspace{-0.7em}
\end{equation}
Finally, a max pooling layer over $e'_i$ followed by a linear layer produces the predicted document class $y$. The model is trained with cross entropy loss.

Additionally, the released L\&L implementation has a bug where attention scores and marginals are not masked correctly in the matrix-tree computation, which we correct.

\smallskip
\noindent \textbf{Sentence order discrimination} This model is identical, except for task-specific changes. The goal of this synthetic task, proposed by \citet{Barzilay:2008}, is to capture discourse coherence. A negative class is created by generating 
random permutations of a text's original sentence ordering (the positive class). A coherence score is produced for each positive and negative example, with the intuition that the originally ordered text will be more coherent than the jumbled version. Because we compare two examples at a time (original and permuted order), we modify the model to handle paired inputs and replace cross-entropy loss with a max-margin ranking loss.

\section{Experiments}
We evaluate the model on four text classification tasks and one sentence order discrimination task. 

\subsection{Datasets}
Details and statistics are included in Appendix \ref{sec:appendix}.\footnote{Of the document-level datasets used in L\&L (SNLI was sentence-level), we omit IMDB and Czech Movies because on IMDB their model did not outperform prior work, and Czech (a language with freer word order than English) highlighted the non-projectivity of their sentence-level trees, which is not the focus of our work.}

\smallskip
\noindent\textbf{Yelp} (in L\&L, 5-way classification) comprises customer reviews from the Yelp Dataset Challenge
(collected by \citet{Tang:2015}). Each review is labeled with a 1 to 5 rating (least to most positive).

\smallskip
\noindent\textbf{Debates} (in L\&L, binary classification) are transcribed debates on Congressional bills from the U.S. House of Representatives (compiled by \citet{Thomas:2006}, preprocessed by \citet{Yogatama:2014}). Each speech is labeled with 1 or 0 indicating whether the speaker voted in favor of or against the bill.   

\smallskip
\noindent\textbf{Writing quality (WQ)} (not in L\&L, binary classification) contains science articles from the New York Times (extracted from \citet{Louis:2013}). Each article is labeled as either `very good' or `typical' to describe its writing quality. While both classes contain  well-written text, \citet{Louis:2013} find features associated with discourse including
sentiment, readability, along with PDTB-style discourse relations are helpful in distinguishing between the two classes.

\smallskip
\noindent\textbf{Writing quality with topic control (WQTC)} (not in L\&L, binary classification) is similar to \textsc{WQ}, but controlled for topic using a topic similarity list included with the \textsc{WQ} source corpus.\footnote{An analysis in section \ref{sec:results} shows the \textsc{WQ}-trained model focuses on lexical items strongly related to the article topic.}  

\smallskip
\noindent\textbf{Wall Street Journal Sentence Order (WSJSO)} (not in L\&L, sentence order discrimination) is the WSJ portion of PTB \cite{Marcus:1993}.

\subsection{Settings}
For each experiment, we train the model four times varying only the random seed for weight initializations. The model is trained for a fixed amount of time, and the model from the timestep with highest development performance is chosen. We report accuracies on the test set, and tree analyses on the development set. Our implementation is built on the L\&L released implementation, with changes as noted in Section \ref{sec:models}. Preprocessing and training details are in Appendix \ref{sec:appendix}.

\subsection{Results}
\label{sec:results}
We report accuracy (as in prior work) in Table \ref{tab:results_full}, and perform two ablations: removing the structured attention at the document level, and removing it at both document and sentence levels. Additionally, we run experiments on the original code \emph{without} the design change or bug fix to confirm our findings are similar (see L\&L(orig) in Table \ref{tab:results_full}).

\medskip
\noindent\textbf{Document-level structured attention does not help.} Structured attention at the sentence level helps performance for all except WQ, where no form of attention helps. However, structured attention at the document level yields mostly negative results, in contrast to the improvements reported in L\&L. In Yelp, WSJSO, and WQ, there is no difference. In Debates, the attention hurts performance. Only in WQTC does the structured attention provide a benefit. While a single training run could produce the improvements seen in L\&L, the results across four runs depict a more accurate picture. When inducing structures, it is particularly important to repeat experiments as the structures can be highly inconsistent due to the noise caused by random initialization \cite{Williams:2018}.   



\begin{figure}[!htb]
\scalebox{0.95}{
\begin{dependency}[theme=simple,hide label,label style={font=\bfseries,thick},edge style={black!60!black,thick}]
  \begin{deptext}
    1 \& 2 \& 3 \& 4 \& 5 \& 6 \& 7 \& 8\& 9\& 10\& 11\& 12\& 13\& 14\& 15\& 16\& 17\\
  \end{deptext}
  \deproot[show label,edge unit distance=1.4ex]{1}{ROOT}
  \depedge[arc angle=30]{2}{1}{}
  \depedge[arc angle=30]{3}{1}{}
  \depedge[arc angle=30]{4}{1}{}
  \depedge[arc angle=30]{5}{1}{}
  \depedge[arc angle=30]{6}{1}{}
  \depedge[arc angle=30]{7}{1}{}
  \depedge[arc angle=30]{8}{1}{}
  \depedge[arc angle=30]{9}{1}{}
  \depedge[arc angle=30]{10}{1}{}
  \depedge[arc angle=30]{11}{1}{}
  \depedge[arc angle=30]{12}{1}{}
  \depedge[arc angle=30]{13}{1}{}
  \depedge[arc angle=30]{14}{1}{}
  \depedge[arc angle=30]{15}{1}{}
  \depedge[arc angle=30]{16}{1}{}
  \depedge[arc angle=30]{17}{1}{}
\end{dependency}}
\small{(1)\textbf{madam speaker, i rise in opposition to h.r. 3283 on both process and policy grounds.}\ldots 
\small{(17)look beyond the majority's smoke and mirrors, and vote against this ill-timed and ill-conceived legislation.}}
\vspace{-0.7em}
\caption{Learned dependency tree from Debates.}
\label{fig:tree2}
\vspace{-1.1em}
\end{figure}


\begin{table}[t]
\small
\centering
\renewcommand{\tabcolsep}{1.3mm}
\scalebox{0.94}{
\begin{tabular}{llllll}
\toprule
                    & Yelp & Debates & WQ & WQTC & WSJSO  \\
\midrule
tree height     & 2.049 &2.751 &2.909 &4.035 &2.288\\
prop. of leaf nodes  & 0.825 & 0.849 &0.958 &0.931 &0.892\\
norm. arc length & 0.433 &0.397 &0.420 &0.396 &0.426\\
\% vacuous trees & 73\% &38\% &42\% &14\%  &100\%\\
\bottomrule
\end{tabular}
}
\vspace{-0.7em}
\caption{Statistics for learned trees averaged across four runs (similar results without the design change or bug fix are in the Appendix Table \ref{tab:trees_full_orig}). See Table \ref{tab:results_deeper} for gold statistics on WQTC.}
\label{tab:trees_full}
\vspace{-0.7em}
\end{table}

\medskip
\noindent\textbf{Trees do not learn discourse.} Although document  level structured attention provides little benefit in performance, we probe whether the model could still be learning some discourse. We visually inspect the learned $\mathbf{f_{best}}$ trees and in Table \ref{tab:trees_full} we report statistics on them (see Appendix Table \ref{tab:trees_full_orig} for similar results with the original code).

The visual inspection (Figure \ref{fig:tree2}) reveals shallow trees (also reported in L\&L), but furthermore the trees have little to no structure.\footnote{While shallow trees are expected in PDTB-style discourse, even these trees would exhibit meaningful structure between adjacent sentences, which is entirely absent here. 
} We observe an interesting pattern where the model picks one of the first two or last two sentences as the root, and all other sentences are children of that node. We label these trees as `vacuous' and the strength of this pattern is  reflected in the tree statistics (Table \ref{tab:trees_full}). The height of trees is small, showing the trees are shallow. The proportion of leaf nodes is high, that is, most nodes have no children. Finally, 
the normalized arc length is high, where nodes that are  halfway into the document still connect to the root. 

We further probe the root sentence, as the model places so much attention on it. We hypothesize the root sentence has strong lexical cues for the task, suggesting the model is instead attending to particular words. In Yelp, reviewers often start or end with a sentiment-laden sentence summarizing their rating. In Debates, speakers begin or end their speech by stating their stance on the bill. In WQ and WQTC, the interpretation of the root is less clear. In WSJSO, we find the root is always the first sentence of the correctly ordered document, which is reasonable and commonly attested in a discourse tree, but the remainder of the vacuous tree is entirely implausible.

To confirm our suspicion that the root sentence is lexically marked, we measure the association between words appearing in the root sentence and those elsewhere by calculating their positive pointwise mutual information scores (Table \ref{tab:latent_ppmi}). 

\begin{table}[t!]
\small
     \begin{center}
     \begin{tabular}{lp{5.5cm}}
\toprule
Yelp                & uuu, sterne, star, rating, deduct, 0, edit, underwhelmed, update, allgemein \\ 
Debates & oppose, republican, majority, thank, gentleman, leadership, california, measure, president, vote    \\ 
WQ     & valley, mp3, firm, capital, universal, venture, silicon, analyst, capitalist, street        \\ \bottomrule
\end{tabular}
\vspace{-0.7em}
\caption{Top 10 words most associated with the root sentence (measured with PPMI).}
      \label{tab:latent_ppmi}
      \end{center}
      \vspace{-1.7em}
      \end{table}
      
In Yelp, we find root words often express sentiment and explicitly mention the number of stars given (`sterne' in German, or `uuu' as coined by a particularly prolific Yelper), which are clear indicators of the rating label. For Debates, words express speaker opinion, politeness and stance which are strong markers for the binary voting label. The list for WQ revolves around tech, suggesting the model is learning topics instead of writing quality. Thus, in WQTC we control for topics.


\section{Learning better structure} We next probe whether the structure in L\&L can be improved to be more linguistically appropriate, while still performing well on the end task. Given that structured attention helps only on WQTC and learns vacuous trees less frequently, we focus on this task. We experiment with three modifications. First, we remove the document-level biLSTM since it performs a level of composition that might prevent the attention from learning the true structure. Second, we note equation \ref{eq:children} captures possible children only at one level of the tree, but not possible subtrees. We thus perform an additional level of percolation over the marginals to incorporate the children's children of the tree. That is, after equation \ref{eq:update}, we calculate:
\vspace{-0.5em}
\begin{equation}
c'_i = \sum^{n}_{k=1}a_{ik}e'_{i}\hspace{1pt};\hspace{20pt}e''_i = \tanh(W_r[e'_i, c'_i])
\vspace{-0.3em}
\end{equation}
Third, the max-pooling layer gives the model a way of aggregating over sentences while ignoring the learned structure. Instead, we propose a sum that is weighted by the probability of a given sentence being the root, i.e., using the learned root attention score $a_i^r$: $y_i=\sum_{i=1}^n a_i^r e''_i$.

\begin{table}[t]
\centering
\small
\scalebox{0.94}{
\begin{tabular}{lllllll}
\toprule
                    & Acc & height & leaf & arc &vacuous \\
\midrule
Full          & \textbf{81.11} & 4.035 & 0.931 &0.396 &14\%\\
-biLSTM & 77.80 & 11.51 & 0.769 &0.353 &4\%\\
-biLSTM, +w & 75.57 &7.364 &0.856 &0.359 & 3\% \\
-biLSTM, +p &77.11 &10.430 &0.790 &0.349 &3\% \\
-biLSTM, +4p & \textbf{81.71} & 9.588 & 0.811 & 0.353 & 3\% \\

\midrule
parsed RST &- &25.084 &0.567 &0.063 &0\%\\
\bottomrule
\end{tabular}}
\vspace{-0.7em}
\caption{Mean test accuracy and tree statistics on the WQTC dev set (averaged across four runs).  -biLSTM removes the document-level biLSTM, +w uses the weighted sum, +p performs 1 extra percolation, and +4p does 4 levels of percolation. The last row are (`gold') parsed RST discourse dependency trees.}
\label{tab:results_deeper}
\vspace{-0.8em} 
\end{table}

We include ablations of these modifications and additionally derive RST discourse dependency trees,\footnote{We use the RST parser in \citet{Feng:2014} and follow \citet{Hirao:2013} to derive discourse dependency trees.} collapsing intrasentential nodes, as an approximation to the ground truth.

The results (Table \ref{tab:results_deeper}) show that simply removing the biLSTM produces trees with more structure (deeper trees, fewer leaf nodes, shorter arc lengths, and less vacuous trees), confirming our intuition that it was doing the work for the structured attention. However, it also results in lower performance. Changing the pooling layer from max to  weighted sum both hurts performance and results in shallower trees (though still deeper than Full), which we attribute to this layer still being a pooling function. Introducing an extra level of tree percolation yields better trees but also a drop in performance. Finally, using 4 levels of percolation both reaches the accuracy of Full and retains the more structured trees.\footnote{More than 4 levels caused training to become unstable.} We hypothesize accuracy doesn't surpass Full because this change also introduces extra parameters for the model to learn.


While our results are a step in the right direction, the structures are decidedly not discourse when compared to the parsed RST dependency trees, which are far deeper with far fewer leaf nodes, shorter arcs and no vacuous trees. Importantly, the tree statistics show the structures do not follow the typical right-branching structure in news: the trees are shallow, nodes often connect to the root instead of a more immediate parent, and the vast majority of nodes have no children. In work concurrent to ours, \citet{Liu:2019} proposes a new iterative algorithm for the structured attention (in the same spirit as our extra percolations) and applies it to a transformer-based summarization model. However, even these induced trees are not comparable to RST discourse trees. The induced trees are multi-rooted by design (each root is a summary sentence) which is unusual for RST;\footnote{Less than 25\% of trees in the RST Discourse Treebank \cite{Carlson:2001} have more than 1 root; less than 8\% have more than 2 roots.} their reported tree height and edge agreement with RST trees are low.

\section{Conclusion}
In this paper, we evaluate structured attention in document representations as a proxy for discourse structure. We first find structured attention at the document level is largely unhelpful, and second it instead captures lexical cues resulting in vacuous trees with little structure. 
We propose several principled changes to induce better structures with comparable performance. Nevertheless, calculating statistics on these trees and comparing them to parsed RST trees shows they still contain no meaningful discourse structure. 
We theorize some amount of supervision, such as using ground-truth discourse trees, is needed for guiding and constraining the tree induction.

\section*{Acknowledgments}
We thank the reviewers for insightful feedback. We acknowledge the Texas Advanced Computing Center for grid resources. The first author was supported by the NSF Graduate Research Fellowship Program under Grant No. 2017247409. 

\bibliography{acl2019}
\bibliographystyle{acl_natbib}
\appendix

\section{Appendices}
\label{sec:appendix}
\begin{table*}[t]
\centering
\small
\begin{tabular}{lllllll}
\toprule
                    &         & \multicolumn{4}{c}{Number of documents}  &             \\ \cline{3-6}
Dataset             & Classes & Total & Train & Dev & Test    & Vocab. \\ 
\midrule
Yelp           & 5       & 333K  & 266,522  & 33,333  & 33,317 & 53K        \\
Debates & 2       & 1.5K  & 1,050    & 102 & 374 & 21K        \\
WQ     & 2       & 7.7K  & 6,195    & 775   & 763     & 150K       \\
WQTC     & 2       & 7.8K  & 6,241    & 777         & 794     & 131K       \\
WSJSO     & -       &2.4K    & 1,950 (35,165)         & 247 (4,392)    & 241 (4,383) &49K\\ \bottomrule
\end{tabular}
\caption{Statistics for the datasets used in the text classification and discrimination tasks (calculated after preprocessing). For WSJSO, the number of generated pairs are in parentheses.}
\label{tab:corpora}
\end{table*}
\paragraph{Datasets} Statistics for the datasets are listed in Table \ref{tab:corpora}. 

\begin{table*}[t]
\small
\centering
\renewcommand{\tabcolsep}{1.3mm}
\begin{tabular}{llllll}
\toprule
                    & Yelp & Debates & WQ & WQTC & WSJSO  \\
\midrule
tree height     & 2.049 (2.248) &2.751 (2.444)&2.909 (2.300) &4.035 (2.468) &2.288 (2.368)\\
prop. of leaf nodes  & 0.825 (0.801) & 0.849 (0.869) &0.958 (0.971) &0.931 (0.966) &0.892 (0.888)\\
norm. arc length & 0.433 (0.468) &	0.397 (0.377) & 0.420 (0.377) & 0.396 (0.391)	&0.426 (0.374)\\
\% vacuous trees & 73\% (68\%) &38\% (40\%) &42\% (28\%) &14\% (21\%) &100\% (56\%)\\
\bottomrule
\end{tabular}
\vspace{-0.7em}
\caption{Statistics for the learned trees averaged across four runs on the L\&L(ours) model with comparisons (in parentheses) to results using the original L\&L code without the design change or bug fix.}
\label{tab:trees_full_orig}
\vspace{-0.7em}
\end{table*}

\begin{table*}[t]
\centering
\small
\begin{tabular}{lllllll}
\toprule
                    & Accuracy & tree height & prop. of leaf & parent entr. & norm. arc length & \% vacuous trees \\
\midrule
Full           & 82.49\thinspace$|$\thinspace\textbf{81.11} (0.95) & 4.035 & 0.931 &0.774 &0.396 &14\%\\
-biLSTM & 80.35\thinspace$|$\thinspace\ 77.80 (1.72) & 11.51 & 0.769 &1.876 &0.353 &4\%\\
-biLSTM, +p & 78.72\thinspace$|$\thinspace\ 77.11 (2.18) &10.430 &0.790 &0.349 &0.349 &3\% \\
-biLSTM, +4p & 82.75\thinspace$|$\thinspace\ \textbf{81.71} (0.70) & 9.588 & 0.811 &1.60 & 0.353 & 3\% \\
-biLSTM, +w & 78.46\thinspace$|$\thinspace\ 75.57 (2.52) &7.364 &0.856 &1.307 &0.359 & 3\% \\
-biLSTM, +w, +p & 77.08\thinspace$|$\thinspace\ 74.78 (2.58) &8.747 & 0.826 & 1.519 &0.349 &4\%\\
\midrule
parsed RST &- &25.084 &0.567 & 2.711 &0.063 &0\%\\
\bottomrule
\end{tabular}
\caption{Max\thinspace$|$ mean (standard deviation) test accuracy and tree statistics of the WQTC dev set (averaged across four training runs with different initialization weights). Bolded numbers are within 1 standard deviation of the best performing model. +w uses the weighted sum, +p adds 1 extra level of percolation, +4p adds 4 levels of percolation. The last row are the (`gold') parsed RST discourse dependency trees.}
\label{tab:results_deeper_detail}
\end{table*}

For WQ, the ‘very good’ class was created by \citet{Louis:2013} using as a seed the 63 articles in the New York Times corpus \cite{Sandhaus:2008} deemed to be high-quality writing by a team of expert journalists. The class was then expanded by adding all other science articles in the NYT corpus that were written by the seed authors (4,253 articles). For the ‘typical’ class, science articles by all other authors were included (19,520). Because the data is very imbalanced, we undersample the ‘typical’ class to be the same size as the ‘very good’. We split this data into 80/10/10 for training, development and test, with both classes equally represented in each partition. 

For WQTC, the original dataset authors provide a list of the 10 most topically similar articles for each article.\footnote{\url{http://www.cis.upenn.edu/~nlp/corpora/scinewscorpus.html}} We make use of this list to explicitly sample topically similar documents.

\paragraph{Preprocessing} For Debates and Yelp, we follow the same preprocessing steps as in L\&L, but do not set a minimum frequency threshold when creating the word embeddings. 
For our three datasets, sentences are split and tokenized using Stanford Core NLP.

\paragraph{Training} For all models, we use the Adagrad optimizer with a learning rate of 0.05. For WQ, WQTC, and WSJSO, gradient clipping is performed using the global norm with a ratio of 1.0. The batch size is 32 for all models except WSJSO uses 16. All models are trained for a maximum of 8 hours on a GeForce GTX 1080 Ti card.

\paragraph{Results}  Because our results hinge on multiple runs of experiments each initialized with different random weights, we include here more detailed versions of our results to more accurately illustrate their variability. Table \ref{tab:trees_full_orig} supplements Table \ref{tab:trees_full} with tree statistics from L\&L(orig), the model \emph{without} the design change or bug fix, to illustrate the derived trees on this model are similar. Finally, Table \ref{tab:results_deeper_detail} is a more detailed version of Table \ref{tab:results_deeper}, which additionally includes  maximum accuracy, standard deviation for accuracy, as well as the average parent entropy calculated over the latent trees. 
\end{document}